\documentclass{statsoc}
\usepackage{geometry}
\usepackage{hyperref}
\usepackage{natbib}
\usepackage{mathtools}
\usepackage{xcolor}

\usepackage{amsmath, amsfonts, amsthm,amssymb}
\usepackage{url}
\usepackage{multirow}
\usepackage{bm,bbm}
\usepackage[ruled]{algorithm2e}

 \newcommand{\eps}{\epsilon}

\newcommand{\beq}{\begin{equation}}
\newcommand{\eeq}{\end{equation}}
\newcommand{\beas}{\begin{eqnarray*}}
\newcommand{\eeas}{\end{eqnarray*}}
\newcommand{\bea}{\begin{eqnarray}}
\newcommand{\eea}{\end{eqnarray}}
\newcommand{\bei}{\begin{itemize}}
\newcommand{\eei}{\end{itemize}}

\newcommand{\ben}{\begin{enumerate}}
\newcommand{\een}{\end{enumerate}}
\newcommand{\argmin}{\mathop{\rm arg\min}}
\newcommand{\argmax}{\mathop{\rm arg\max}}

\def\bx{\bm{x}}

\newtheorem{corollary}{Corollary}
\newtheorem{Proposition}{Proposition}

\newtheorem{lemma}{Lemma}
\newtheorem{definition}{Definition}

\newtheorem{theorem}{Theorem}
\newtheorem{example}{Example}

\newtheorem{remark}{Remark}

\newtheorem{condition}{Condition}[section]

\newcommand{\bbeta}{\bm{\beta}}
\newcommand{\R}{\mathbb{R}}
\newcommand{\E}{{\mathbb{E}}}

\newcommand{\MM}{{\mathcal{M}}}

\newcommand{\Var}{{\rm Var}}

  \def\P{\mathbb{P}}

  \def\calF{\mathcal{F}}
    \def\calE{\mathcal{E}}
    \def\E{\mathbbm{E}}
    
  \def\eps{\epsilon}

    \def\lam{\lambda}
\def\Sig{\bm \Sigma}

\def\ERM{\textup{ERM}}
\def\MM{\textup{MM}}
\def\FAIRM{\textup{FAIRM}}
\def\CMI{\textup{CMI}}

\title[FAIRM]{FAIRM: Learning invariant representations for algorithmic fairness and  domain generalization with minimax optimality}

\author{Sai Li\thanks{
Supported in part by the National Natural Science Foundation of China (grant no. 12201630).}}
\address{Institute of Statistics and Big Data, Renmin University of China, China.}
\email{saili@ruc.edu.cn}

\author[Li and Zhang]{Linjun Zhang\thanks{Corresponding author.
Supported by NSF DMS-2015378, NSF CAREER DMS-2340241.}}
\address{Department of Statistics, Rutgers University, US}
\email{linjun.zhang@rutgers.edu}
\begin{document}

\begin{abstract}
Machine learning methods often assume that the test data have the same distribution as the training data. However, this assumption may not hold due to multiple levels of heterogeneity in applications, raising issues in algorithmic fairness and domain generalization. In this work, we address the problem of fair and generalizable machine learning by invariant principles. We propose a training environment-based oracle, FAIRM, which has desirable fairness and domain generalization properties under a diversity-type condition. We then provide an empirical FAIRM with finite-sample theoretical guarantees under weak distributional assumptions. We then develop efficient algorithms to realize FAIRM in linear models and demonstrate the nonasymptotic performance with minimax optimality.  We evaluate our method in numerical experiments with synthetic data and MNIST data and show that it outperforms its counterparts.
\end{abstract}

\section{Introduction}

The conventional setting of machine learning is that the test data have identical distributions as the training data, under which assumptions many algorithms have been developed for different purposes. However, real-world datasets often exhibit heterogeneity, leading to distribution shifts in both the training and test data. This paper focuses on the setting where the training samples are generated from multiple different distributions and predictions are made based on the heterogeneous data. For example, in medical applications, data collected from various hospitals may differ significantly due to factors such as geographical location, which can influence patient demographics, economic status, and lifestyle habits. Training a prediction algorithm on such heterogeneous datasets and applying it to a new hospital may lead to unexpected performance, as the test distribution may be markedly different from any training distribution.
This scenario exemplifies the out-of-distribution (OOD) setting, which is a critical challenge regarding the robustness of modern machine learning.

This work is dedicated to exploring reliable prediction algorithms suited for the OOD setting. We focus on two fundamental aspects of prediction algorithms: domain generalization and algorithmic fairness. Domain generalization refers to the ability of an algorithm to adapt and maintain accuracy across different distributions. Algorithmic fairness, on the other hand, addresses the potential biases in prediction, ensuring that the algorithm's output is equitable across diverse subpopulations. By tackling these aspects, we aim to contribute to the development of machine learning models that are both effective and responsible, particularly in settings where traditional assumptions about data homogeneity no longer hold.


\subsection{Problem description}
We consider the setting where the data are generated from multiple underlying environments or subpopulations. Denote the set of all the environments as $\calE_{all}$ and the set of training environments as $\calE_{tr}$ with $\calE_{tr}\subseteq \calE_{all}$. Suppose that the test data are generated from some environment $e^*\in\calE_{all}$ but the test data and its environment $e^*$ are unseen in the training phase. Our main focus is the OOD setting where $e^*\notin\calE_{tr}$, i.e., the test environment is not represented in the training domains. 
In contrast, conventional supervised learning scenarios assume that $e^*\in\calE_{tr}$ and $e^*$ is known in the training phase. 

For each environment $e\in\calE_{all}$, we denote the features and response variable as $\bx^e\in\R^p$ and $y^e\in\R$, respectively.  Their joint distribution is represented as $((\bx^e)^\top ,y^e)\sim P^e$, where $P^e \neq P^{e'}$ in general for $e \neq e' \in \mathcal{E}_{all}$. For each environment $e \in \mathcal{E}_{tr}$, we observe $n_e$ independent samples, denoted by $\{\bx_i^e,y_i^e\}_{i=1}^{n_e}$. If $e \notin \mathcal{E}_{tr}$, then $n_e = 0$.


\textbf{Domain Generalization:} This aspect concerns prediction accuracy in the OOD setting. We quantify domain generalization accuracy using\\ $\max_{e\in\calE_{all}}\E[(y^{e}-f(\bx^{e}))^2]$ for a given prediction function $f(\cdot)$. This metric assesses the ability of $f(\cdot)$ to generalize from $\calE_{tr}$ to any domains in $\calE_{all}$.

 
 \textbf{Algorithmic Fairness:} This aspect concerns how an algorithm performs differentially with respect to sensitive (categorical) attributes, e.g., gender and race. By defining environments based on sensitive attributes, the goal of algorithmic fairness aligns with achieving comparable performance across different environments. To quantify the fairness of a prediction rule, we consider multi-calibration \citep{hebert2018multicalibration}.

\begin{definition}[Multi-calibration (adapted from \citet{hebert2018multicalibration})]
\label{def-mc}
For a set of environments $\mathcal{E}_{all}$, we say a prediction rule $f(\cdot)$ is $\alpha$ multi-calibrated with respect to $\mathcal E_{all}$  if 
$$
\sup_{e\in\calE_{all}}\sup_{\nu\in\R}|\E_{(\bx^{\top},y)\sim P^e}[y-f(\bx)\mid f(\bx)=\nu]|\le\alpha.
$$
\end{definition}
In simpler terms, a small $\alpha$ indicates that the estimator exhibits comparable prediction biases across all the environments or subpopulations. \citet{hebert2018multicalibration} define environments as a collection of subsets $\mathcal A=\{A_1, ..., A_K\}$ where $A_k\subseteq \mathcal X$ and the multi-calibration property of $\hat{f}$ is defined with respect to $\mathcal{A}$.
Definition \ref{def-mc} can be rewritten in the same fashion if we consider $\mathcal X\times \calE_{all}$ as the feature space and define the collection of sets by  $\mathcal A=\{(\bm x^{\top},e)\in\mathcal X\times\calE_{all}: e\in\calE_{all}\}$. 
It is important to note that protected attributes need not be the explicit inputs to the prediction model $f(\bm{x})$, a critical consideration in sectors where using such attributes is legally restricted  
 \citep{fremont2016race,zhang2018assessing,awasthi2021evaluating}.
This fairness notion, multi-calibration, has been widely used in recent years as it does not have a fairness-accuracy trade-off \citep{kleinberg2016inherent, hebert2018multicalibration,kearns2019empirical,burhanpurkar2021scaffolding,deng2023happymap}. While the concept of multi-calibration is traditionally investigated in settings where all subpopulations are well-represented in the training data, we extend it to the more challenging OOD setting, thereby broadening its scope and applicability.

To address the above-mentioned two problems, we recognize that one common cause of distributional shifts in the training and test data is the existence of unmeasured confounders. The unmeasured confounders do not affect the outcome directly but are associated with observed features, which can lead to spurious associations between the confounders and the outcome. Incorporating these spurious associations in prediction models can be harmful to OOD prediction and algorithmic fairness. For instance, consider the face image classification dataset CelebA \citep{liu2015faceattributes} that classifies whether the hair color in a face image is blonde or dark, in which most training images of
blonde hair are female and dark hair are male. When trained on such a dataset, models are found to easily rely on the spurious features (gender-related features) rather than the truly responsible features (hair-related features) 
to make predictions. It can lead to biased model prediction \citep{geirhos2020shortcut}, i.e., images of females are more likely to be classified as having blonde hair. If test images consist of males with blonde hair and females with dark hair, the misclassification error of such a prediction rule can be large. 
To avoid biased prediction, we are motivated to predict with hair-related features as their relationship with the outcome is ``invariant'' in diverse subpopulations. To do this, we need to distinguish invariant features (e.g., hair) from spurious features (e.g., gender).

In summary, our goal is to develop algorithms based on invariance principles that achieve both reliable domain generalization and multi-calibration in the OOD setting. The key distinction between these two goals is that multi-calibration evaluates an algorithm's bias in each domain, while domain generalization focuses on the overall prediction error, a combined measure of bias and variance. 
\subsection{Related works}

It is known that the causal rules generalize well under linear structural equation models (SEMs) because the causal effects are invariant under interventions \citep{pearl2009causality}. However, the causal relationships are hard to learn based on the observational data due to the existence of unknown confounders.
To identify the causal features which are directly responsible for the outcome, the state-of-the-art approach focuses on learning invariant representations. 
The recent work of \citet{peters2016causal} and \citet{buhlmann2020invariance} establish that a subset of causal features can be learned by identifying the invariant features in linear SEMs. 
Along this line of research,  \citet{pfister2019invariant} study selecting causal features with sequential data based on the invariant principle.  \citet{chen2021domain} propose new estimands for domain generalization with theoretical guarantees under linear SEMs. \citet{pfister2021stabilizing} investigate so-called stable blankets for domain generalization but the proposed algorithm does not have theoretical guarantees.

Without the assumptions of SEMs, the causality of a prediction rule is hard to justify and it is of interest to directly learn invariant representations for the purpose of domain generalization. \citet{rojas2018invariant} develop the invariant risk minimization (IRM) framework, aiming to discover invariant representations across multiple training environments while excluding spurious features. However, empirical studies by \citet{kamath2021does} and \citet{choe2020empirical} suggest that the sample-based implementation of IRM may not consistently capture the intended invariance. \citet{rosenfeld2021risks} delve into the risks associated with IRM in linear discriminant analysis models, particularly under infinite sample conditions, and explores scenarios where the IRM framework might falter in both linear and nonlinear models. \citet{fan2023environment} study an invariant least square approach for domain generalization under infinite sample conditions.

Another prominent approach for robust domain generalization is the Maximin estimator introduced by \citet{meinshausen2015maximin}. In the machine learning context, this methodology aligns with what is known as group distributionally robust optimization (group DRO). The group DRO concept has been extensively explored and expanded upon in several studies, including \citet{namkoong2016stochastic}, \citet{sagawa2019distributionally}, \citet{martinez2020minimax}, and \citet{diana2021minimax}. 
Additionally, self-training and the common subspace approach have gained traction as popular methods for domain generalization. For instance, \citet{kumar2020understanding} delve into the theoretical properties of self-training, particularly in contexts characterized by gradual shifts. On another front, \citet{pan2010domain} propose an innovative domain invariant projection approach by extracting the information that is invariant across the source and target domains. 




\subsection{Our contributions}
We propose a new invariant risk minimization framework, termed FAIRM, for reliable OOD prediction. We show that the  FAIRM, an oracle algorithm defined at the population level based on training domains, has desirable invariance properties under a diversity-type condition.
It outperforms empirical risk minimization and Maximin methods in terms of minimax domain generalization risks and multi-calibration errors. 
To the best of our knowledge, FAIRM is the first training environment-based oracle that has guaranteed invariance. Such an invariance turns out to lead to small domain generalization error and algorithmic fairness at the same time.

Further, FAIRM can be easily realized with finite samples. We propose empirical FAIRM, a generic training paradigm, and provide theoretical guarantees under almost distribution-free assumptions. 
We then adapt empirical FAIRM to high-dimensional linear models and introduce a computationally efficient algorithm. Our method incorporates the strategies of variable selection in multi-task learning and it works for both fixed-dimensional and high-dimensional features. The proposed method is guaranteed to find invariant representations and enjoys minimax optimal domain generalization performance.

\subsection{Organization and Notation}
In Section \ref{sec2}, we introduce the full-information FAIRM and discuss its connections to the existing works. In Section \ref{sec3}, we introduce FAIRM and prove its desirable properties for domain generalization and multi-calibration. In Section \ref{sec-lm}, we introduce the proposed algorithm in linear models and establish its theoretical performance. In Section \ref{sec-simu}, we evaluate the performance of our proposal in multiple numerical studies. Section \ref{sec-diss} discusses potential extensions beyond the linear models and concludes the paper.

\section{Full-information invariant oracle and its properties}
\label{sec2}

In this section, we propose a new invariant oracle estimand for domain generalization and fairness purposes. This oracle estimand is defined at the population level based on $\calE_{all}$. It serves as the full-information benchmark for algorithms based on training domains. We introduce its definition in Section \ref{sec2-1} and study its invariance and fairness properties in Section \ref{sec2-2}. 

\subsection{Full-information FAIRM}
\label{sec2-1}
We consider the prediction function of the form $w\circ\Phi$, where $\Phi:\mathcal{X}\rightarrow \mathcal{H}$ denotes a data representation and $w:\mathcal{H}\rightarrow\mathcal{Y}$ denotes a prediction rule based on $\Phi$.  Let $e_i$ denote the environment that $(\bx_i^{\top},y_i)$ belongs to and $\alpha^*_{e}=\P(e_i=e)$, $e\in\calE_{all}$. 
We first introduce the \textbf{full-info FAIRM} (\textbf{F}iltration-\textbf{A}ssisted \textbf{I}nvariant \textbf{R}isk \textbf{M}inimization based on full information)  
  \begin{align}\label{full-info}
&(w^*, \Phi^*)=\argmin_{w\in\mathcal F_{w},\Phi\in\mathcal F_{\Phi}} \sum_{e\in \calE_{all}}\alpha_e^*\E[(y^e-w\circ\Phi(\bx^e))^2]~ \\
&\text{subject to}\left\{\begin{array}{ll}
\max\limits_{e,e'\in\mathcal{E}_{all}}\sup\limits_{u\in\mathcal{F}_w}|\E[ \langle u\circ\Phi(\bx^e),y^e\rangle]-\E[ \langle u\circ\Phi(\bx^{e'}),y^{e'}\rangle]|=0\nonumber\\
\max
\limits_{e,e'\in\mathcal{E}_{all}}\sup\limits_{u\in\mathcal{F}_w}|\E[\{u\circ\Phi(\bx^e)\}^2]-\E[\{u\circ\Phi(\bx^{e'})\}^2]|=0,\nonumber
\end{array}\right.
\end{align} 
where $\mathcal{F}_{\Phi}$ is a pre-specified functional class for data representation $\Phi$ and $\mathcal{F}_{w}$ is the functional class for $w$. For example, in linear regression models, $\mathcal F_{\Phi}$ is the class of all subsets of features, and $\mathcal F_{w}$ is the space of linear transformations. In fully connected neural networks,   $\mathcal F_{\Phi}$ is the class of functions that represent neural networks from the first layer to the second last layers and $\mathcal F_{w}$ is the space of linear transformations, i.e., the last layer in a neural network.  
 The first constraint in (\ref{full-info}) demands invariance in the covariance between $y^e$ and $u\circ\Phi(\bm x^e)$ for all $u\in\mathcal{F}_w$. The second constraint ensures invariance in the second moment of $u\circ\Phi(\bm x^e)$ across all $u\in\mathcal{F}_w$.  As we will show in later sections, these two constraints jointly guarantee reliable prediction performance in all the environments. 
While constraints are uniformly applied over $\mathcal{F}_w$, it may not be necessary to examine every function in $\mathcal{F}_w$ in some typical model classes. For example, in Section \ref{sec-lm}, we consider linear prediction rules where the supremum can be equivalently represented by specific functions.

A notable feature of (\ref{full-info}) is that both constraints are independent of the function $w(\cdot)$. This independence enables us to transform the constrained optimization problem into a filter-then-optimize approach, which can be realized more efficiently. Specifically, let $\mathcal{I}^*_{\Phi}\subseteq \mathcal{F}_{\Phi}$ denote the set of feasible solutions to (\ref{full-info}). Once $\mathcal{I}^*_{\Phi}$ is identified, we can solve for $(w^*, \Phi^*)$ using the following optimization:
  \begin{align}
(w^*, \Phi^*)=&\argmin_{w\in\mathcal F_{w},\Phi\in\mathcal{I}^*_{\Phi}} \sum_{e\in \calE_{all}}\alpha_e^*\E[(y^e-w\circ\Phi(\bx^e))^2].\label{rewrite1}
\end{align} 


\begin{remark}[Feasibility of full-info FAIRM]
A sufficient condition for the feasibility of (\ref{full-info}) is $\E[y^e]=\mu_0\in\R$ for all $e\in\calE_{all}$. In this case, $\Phi(\bx^e)=1$ is feasible to (\ref{full-info}). Another sufficient condition is that $\E[y^e|\Phi'(\bx^e)=\bm\phi]=w(\bm\phi)$ for some $\Phi'\in\mathcal{F}_{\Phi}$ satisfying the second constraint of (\ref{full-info}). 
\end{remark}
The condition of equal $\E[y^e]$ for all $e\in\calE_{all}$ is commonly used in the literature of domain adaptations; see \cite{redko2020survey} and the references therein. The second sufficient condition is related to the conditional mean invariance defined in Assumption 1 of \citet{peters2016causal}.
The feasibility assumption can be diagnosed by checking whether the estimated feasible set is empty or not. 

In contrast to existing works, the full-info FAIRM (\ref{full-info}) prioritizes prediction accuracy rather than seeking a causal prediction rule as illustrated in the next example.
\begin{example}
\label{ex1}
Consider $y^e=x_1^e+x_2^e+\eps^e$ where $(x_1^e,x_2^e,\eps^e)\sim N(0, I_3)$ for $e\in\calE_{all}$. Then $0, x^e_1, x^e_2$, and $(x_1^e,x_2^e)$ are all invariant representations based on the Assumption 1 of \citet{peters2016causal} and  they are also elements in $\mathcal{I}^*_{\Phi}$ for $\mathcal{F}_{\Phi}=\{\bx_S:S\subseteq [p]\}$. 
\end{example}
As shown in Example \ref{ex1}, there can be multiple invariant representations in certain scenarios {and some of them can have low prediction power.}
 \citet{peters2016causal} and \citet{buhlmann2020invariance} show that the intersection of invariant sets identifies a subset of causal predictors in linear Gaussian SEMs.  However, as Example \ref{ex1} illustrates, the intersected set can yield a narrow set of covariates with limited predictive capability. \citet{fan2023environment} assumes the uniqueness of the invariant representations, which is violated in this simple example. {\color{black}In contrast, full-info FAIRM looks for the invariant prediction rule that has the best prediction accuracy when multiple invariant representations exist.}


\subsection{OOD performance of full-info FAIRM}
\label{sec2-2}
In this subsection, we show that the $w^{*}\circ\Phi^{*}$ defined by \eqref{full-info}  has desirable domain generalization properties and satisfies the algorithmic fairness criterion. 
We consider the following distribution space 
\begin{align*}
  \Xi(\rho,\sigma^2)&=\left\{\mathcal{P}=\{P^e\}_{e\in\calE_{all}}:~\mathcal{I}^*_{\Phi}\neq \emptyset,~ \text{(\ref{cond-recover}) holds},~\max_{e\in\calE_{all}}\Var(y^e|\bx^e)\leq \sigma^2,\right.\\
 & \left.~~\quad\max_{e\in\calE_{all}}\E[\{\E[y^e|\bx^e]-\omega^*\circ\Phi^*(\bx^e)\}^2]\leq \rho\right\},
\end{align*}
where $\Var(y^e|\bx^e)$ denotes the variance of $y^e$ conditional on $\bx^e$.
The parameter $\rho$ characterizes the strengths of spurious correlations caused by variant features.
For a prediction function $f(\cdot)$, denote its prediction risk under environment $e$ as $R^e(f)=\mathbb{E}_{(\bm x^\top ,y)\sim P^e}[(y-f(\bx))^2]$, where $e\in\mathcal{E}_{all}$ and $(\bm x^\top ,y)$ is independent of $f(\cdot)$. 

\begin{Proposition}[OOD performance of full-info FAIRM]
\label{prop1}
\[
\sup_{\mathcal{P}\in   \Xi(\rho,\sigma^2)}\max_{e\in\calE_{all}}R^e(w^*\circ\Phi^*)=\inf_{f\in\mathcal{F}_{w}\circ\mathcal{F}_{\Phi}}\sup_{\mathcal{P}\in   \Xi(\rho,\sigma^2)}\max_{e\in\calE_{all}}R^e(f).
\]
If it further holds that $\E[y^e|\Phi(\bx^e)=\bm\phi]=w(\bm\phi)\in\calF_{w}$ for all $\Phi\in\mathcal{I}^*_{\Phi}$, then 
\begin{equation}\label{inv-def}
   \sup_{v\in\R}|\E[y^e|w^*\circ\Phi^*(\bx^e)=v]-v|=0,~\forall~e\in\calE_{all}.
\end{equation}
\end{Proposition}
The first part of Proposition \ref{prop1} shows that the full-info FAIRM is minimax optimal for domain generalization in the parameter space $\Xi(\rho,\sigma^2)$.
The second part implies that $w^*\circ\Phi^*$ achieves perfect multi-calibration if the conditional mean functions belong to $\mathcal{F}_{w}$. 
In Proposition A.1 in the supplement, we also show that $\Phi^*$ qualifies as an invariant representation as defined in \citet{arjovsky2019invariant} and $w^*\circ\Phi^*$ is an invariant prediction rule with respect to the squared loss. To sum up, full-info FAIRM has desirable OOD prediction performance and we will investigate its realization in the following sections.

\section{Training environment-based invariant framework}
\label{sec3}
The full-info FAIRM (\ref{full-info}) is defined based on all environments, while some of which are unseen in the training phase. This section introduces the training paradigm, FAIRM, in Section \ref{sec3-1}, compares FAIRM with current leading methods for OOD prediction in Section \ref{sec3-2}, and presents an empirical version of FAIRM along with theoretical guarantees in Section \ref{sec3-3}.

\subsection{A training environment-based invariant oracle}
\label{sec3-1}
We propose the following training paradigm, termed FAIRM (Filtration-Assisted Invariant Risk Minimization based on training environments), which is designed exclusively on training environments. 
It is defined as
    \begin{align}
    \label{fairm-oracle}
       & (w^{(\FAIRM)},\Phi^{(\FAIRM)})=\argmin_{w\in\mathcal{F}_{w},\Phi\in\mathcal{F}_{\Phi}}\sum_{e\in \calE_{tr}}\alpha_e\E[(y^e-w\circ\Phi(\bx^e))^2]\\
        &\text{subject to} \left\{\begin{array}{ll}
\max\limits_{e,e'\in\calE_{tr}} \sup\limits_{u\in\mathcal{F}_{w}}|\E[ \langle u\circ\Phi(\bx^e),y^e\rangle]-\E[ \langle u\circ\Phi(\bx^{e'}),y^{e'}\rangle]|=0\\
\max\limits_{e,e'\in\calE_{tr}}\sup\limits_{u\in\mathcal{F}_{w}}|\E[\{u\circ\Phi(\bx^e)\}^2]-\E[\{u\circ\Phi(\bx^{e'})\}^2]|=0
        \end{array}\nonumber\right.
\end{align}
where $\alpha_e=n_e/N_{tr}$ for $e\in\calE_{tr}$. The objective function is designed to maximize the prediction accuracy of $w^{(\FAIRM)}\circ\Phi^{(\FAIRM)}$ and the constraints adapt those in (\ref{full-info}) to training environments. 

 In the following, we show that FAIRM can recover its full-information benchmark (\ref{full-info}) under a diversity-type condition.

\begin{Proposition}[FAIRM recovers the full-information benchmark]
\label{prop-recover}
 Assume that $\forall \Phi\in\mathcal{F}_{\Phi}\setminus\mathcal{I}^*_{\Phi}$,
 \begin{align}
   & \max_{e,e'\in\mathcal{E}_{tr}} \sup_{u\in\mathcal{F}_{w}}|\E[\langle u\circ\Phi(\bx^e),y^e\rangle]-\E[\langle u\circ\Phi(\bx^{e'}),y^{e'}\rangle]|>0~~\nonumber\\
     \text{or}~ &   \max_{e,e'\in\mathcal{E}_{tr}}\sup_{u\in\mathcal{F}_{w}}|\E[\{ u\circ\Phi(\bx^e)\}^2]-\E[\{ u\circ\Phi(\bx^{e'})\}^2]|>0.    \label{cond-recover}
\end{align}

Then the feasible set of (\ref{fairm-oracle}) equals $\mathcal{I}^*_{\Phi}$. If $\mathcal{I}^*_{\Phi}\neq \emptyset$, then $w^{(\FAIRM)}\circ\Phi^{(\FAIRM)}=w^{*}\circ \Phi^{*}$.
\end{Proposition}

Proposition \ref{prop-recover} indicates that under the diversity-type condition (\ref{cond-recover}), $w^{(\FAIRM)}\circ\Phi^{(\FAIRM)}$ can recover the full-information benchmark. This condition requires that if either $\mathbb{E}[\langle y^e,u\circ\Phi(\bx^e)\rangle]$ or $\mathbb{E}[\{ u\circ\Phi(\bx^e)\}^2]$ is diverse and variant across $\calE_{all}$, such variability should be detectable within $\calE_{tr}$. This condition are more likely to hold as training environments become more diverse. 

The diversity-type condition (\ref{cond-recover}) highlights the fundamental difference between FAIRM and other frameworks for multi-task learning. Many existing works for multi-task learning make similarity assumptions on the model parameters \citep{tripuraneni2021provable,li2022transfer}. When the model parameters are sufficiently similar among different domains, knowledge can be transferred from training domains to other domains. In contrast, condition (\ref{cond-recover}) can be seen as a ``dissimilarity" criterion among training environments, facilitating the detection of unknown confounders. Furthermore, FAIRM assumes only the invariance of the relationship between $y^e$ and $\Phi^*(\bx^e)$, without specifying how variant features are related to the outcome, making it less assumption-dependent compared to other frameworks \citep{peters2016causal,fan2023environment}. 


 \subsection{Comparison to the prior arts}
 \label{sec3-2}
 
 This subsection compares FAIRM with Empirical Risk Minimization (ERM), Maximin (MM), a method that minimizes the worst-group accuracy, and the  invariance framework adapted from \citet{peters2016causal}.  
 
We first define ERM and MM with respect to the functional classes $\mathcal{F}_w$ and $\mathcal{F}_\Phi$.
\begin{align*}
(w^{(\ERM)},  \Phi^{(\ERM)})&=\argmin_{w\in\mathcal F_{w},\Phi\in\mathcal{F}_{\Phi}} \sum_{e\in \calE_{tr}}\alpha_e\E[(y^e-w\circ\Phi(\bx^e))^2]\\
(w^{(\MM)},\Phi^{(\MM)})&=\argmin_{w\in\mathcal{F}_{w},\Phi\in\mathcal{F}_{\Phi}} \max_{e\in\calE_{tr}} \E[(y^e-w\circ \Phi(\bx^e))^2].
\end{align*}
We see that FAIRM, ERM, and MM are all training-environment-based oracles. 
We also consider the conditional mean-based invariance (CMI) framework adapted from \citet{peters2016causal}, which is defined as:
 \begin{align}
    \label{f-irm-minus}
        (w^{(\CMI)},\Phi^{(\CMI)})=& \argmin_{w\in\mathcal{F}_{w},\Phi\in\mathcal{F}_{\Phi}}\sum_{e\in \calE_{tr}}\alpha_e\E[(y^e-w\circ\Phi(\bx^e))^2]\\
        &\text{subject to} ~~\E[  y^e-w\circ\Phi(\bx^e)|\Phi(\bx^e)]=0~\forall e\in\mathcal{E}_{all}.\nonumber
        \end{align}
        The constraint in (\ref{f-irm-minus}) is defined based on all the environments, ensuring that all feasible solutions are invariant conditional mean functions. The oracle prediction rule $w^{(\CMI)}\circ\Phi^{(\CMI)}$ is defined as the one with the smallest training error among them.
\begin{remark}
\label{re1}
The comparison of $w^{(\CMI)}\circ\Phi^{(\CMI)}$ and $w^{(\FAIRM)}\circ\Phi^{(\FAIRM)}$ should highlight the importance of FAIRM constraints. 
In the following Propositions~\ref{prop: dg} and \ref{prop: fairness}, we will compare them under the diversity condition (\ref{cond-recover}), which guarantees that $\Phi^{(\FAIRM)}\in\mathcal{I}^*_{\Phi}$. Under these conditions, $w^{(\FAIRM)}\circ\Phi^{(\FAIRM)}$ can be equivalently written as
\begin{align*}
     (w^{(\FAIRM)},\Phi^{(\FAIRM)})&= \argmin_{w\in\mathcal{F}_{w},\Phi\in\mathcal{F}_{\Phi}}\sum_{e\in \calE_{tr}}\alpha_e\E[(y^e-w\circ\Phi(\bx^e))^2]\\
    &\quad \text{subject to}~\Phi\in\mathcal{I}^*_{\Phi}.
\end{align*}
Comparing the above formulation with (\ref{f-irm-minus}), we see the only difference between the two methods is in the constraints and we will show that the FAIRM leads to better domain generalization and fairness.
\end{remark}

In the next proposition, we compare FAIRM with ERM, MM, and CMI estimators for domain generalization. 
\begin{Proposition}[Risks for domain generalization]
\label{prop: dg}
Assume that $\mathcal{F}_{\Phi}\supseteq \{\bm{x}_S:S\subseteq[p]\}$ and ${\mathcal{F}}_{w}\supseteq\{w:w(\bm{\phi})=\bm{\phi}^\top \bm{b}\}$.
Then there exists some positive constant $c$ such that
{\small
\begin{align*}
&\sup_{\mathcal{P}\in   \Xi(\rho,\sigma^2)}\max_{e\in\calE_{all}}R^e(w^{(\ERM)}\circ\Phi^{(\ERM)})\geq \sup_{\mathcal{P}\in    \Xi(\rho,\sigma^2)}\max_{e\in\calE_{all}}R^e(w^{(\FAIRM)}\circ\Phi^{(\FAIRM)})+ c\rho\\
&\sup_{\mathcal{P}\in   \Xi(\rho,\sigma^2)}\max_{e\in\calE_{all}}R^e(w^{(\MM)}\circ\Phi^{(\MM)})\geq \sup_{\mathcal{P}\in    \Xi(\rho,\sigma^2)}\max_{e\in\calE_{all}} R^e(w^{(\FAIRM)}\circ\Phi^{(\FAIRM)})+ c\rho\\
& \sup_{\mathcal{P}\in   \Xi(\rho,\sigma^2)}\max_{e\in\calE_{all}}R^e(w^{(\CMI)}\circ\Phi^{(\CMI)})\geq\sup_{\mathcal{P}\in   \Xi(\rho,\sigma^2)}\max_{e\in\calE_{all}}R^e(w^{(\FAIRM)}\circ\Phi^{(\FAIRM)})+c\rho.
\end{align*}
}
\end{Proposition}
Proposition \ref{prop: dg} assumes that the functional spaces contain linear prediction rules. It demonstrates that FAIRM has the smallest minimax risk in $\Xi(\rho,\sigma^2)$ for domain generalization in comparison to ERM, MM, and CMI. The larger errors of ERM and MM are due to their dependence on spurious features. When these variant features have opposite correlations with the outcome in the training and test environments, the prediction rule learned from training domains can have large prediction errors in test domains.
The larger minimax risk of $w^{(\CMI)}\circ\Phi^{(\CMI)}$ is due to the different forms of invariant representations as we discussed in Remark \ref{re1}. Especially, FAIRM asks for the invariance in $\E[\{u\circ\Phi(\bx^e)\}^2]$, which is crucial for the minimax risk guarantee. 

We now compare different algorithms regarding their fairness properties.
\begin{Proposition}[Multi-calibration bias]
\label{prop: fairness}
Assume the $\mathcal{F}_{\Phi}\supseteq \{\bx_S:S\subseteq[p]\}$ and ${\mathcal{F}}_{w}\supseteq\{w:w(\bm\phi)=\bm{\phi}^\top \bm{b}\}$. Assume that $\E[y^e|\Phi(\bx^e)=\bm\phi]=w(\bm\phi)\in\calF_{w}$ for all $\Phi\in\mathcal{I}^*_{\Phi}$. Then
there exists some positive constant $c$ such that for any $v\in\R$,
\begin{align*}
&\sup_{\mathcal{P}\in\Xi(\rho,\sigma^2)}\max_{e\in\calE_{all}}|\E[y^e-  v|w^{(\FAIRM)}\circ\Phi^{(\FAIRM)}(\bx^e)=v]|= 0\\
&\sup_{\mathcal{P}\in\Xi(\rho,\sigma^2)}\max_{e\in\calE_{all}}|\E[y^e-  v|w^{(\CMI)}\circ\Phi^{(\CMI)}(\bx^e)=v]|= 0\\
&\sup_{\mathcal{P}\in\Xi(\rho,\sigma^2)}\max_{e\in\calE_{all}}|\E[{y^{e}}-v|w^{(\ERM)}\circ\Phi^{(\ERM)}(\bx^e)=v]|\geq cv
\\
&\sup_{\mathcal{P}\in\Xi(\rho,\sigma^2)}\max_{e\in\calE_{all}}|\E[y^{e}-v|w^{(\MM)}\circ\Phi^{(\MM)}(\bx^e)=v]|\geq cv.
\end{align*}
\end{Proposition}

Proposition \ref{prop: fairness} demonstrates that FAIRM and CMI are both multi-calibrated in the OOD scenario while ERM and Maximin estimators are not. The results reveal the importance of invariant representation in achieving algorithmic fairness. We also show that $w^{(\FAIRM)}\circ\Phi^{(\FAIRM)}$ can achieve counterfactual fairness \citep{kusner2017counterfactual}, another popular fairness notion, in Section C of the supplements.

\subsection{Empirical FAIRM}
\label{sec3-3}
 We propose a direct realization of FAIRM in generic functional spaces and establish its finite sample guarantees under mild conditions. 

Based on the observed data $(X^e,\bm{y}^e)\in\R^{n_e\times (p+1)}$, $e\in\calE_{tr}$, the empirical FAIRM is defined as follows: 
{\small
\begin{align}
\label{e-FAIRM}
       & (\widehat{w}^{(\FAIRM)},\widehat{\Phi}^{(\FAIRM)})= \argmin_{w\in\mathcal{F}_{w},\Phi\in\mathcal{F}_{\Phi}}\sum_{e\in \calE_{tr}}\|\bm{y}^e-w\circ\Phi(X^e)\|_2^2\\
        &\text{subject to} ~\left\{\begin{array}{ll}
 \max\limits_{e,e'\in\calE_{tr}}\sup\limits_{u\in\mathcal{F}_{w}}\big|\frac{1}{n_e} \langle u\circ\Phi(X^e),\bm{y}^e\rangle-\frac{1}{n_{e'}}\langle u\circ\Phi(X^{e'}),\bm{y}^{e'}\rangle\big|\leq \rho_{1,n}\\
 \max\limits_{e,e'\in\calE_{tr}}\sup\limits_{u\in\mathcal{F}_{w}}\big|\frac{1}{n_e}\sum\limits_{i=1}^{n_e}\{u\circ\Phi(\bx_i^e)\}^2-\frac{1}{n_{e'}}\sum\limits_{i=1}^{n_{e'}}\{u\circ\Phi(\bx_i^{e'})\}^2\big|\leq \rho_{2,n},
        \end{array}\nonumber\right.
\end{align}
}
where $\rho_{1,n}$ and $\rho_{2,n}$ are tuning parameters. They 
  are set based on the upper bounds of the deviations of empirical moments satisfying 
{\small
\begin{align}
&\P\left(\max_{e\in\calE_{tr}}\sup_{u\in\mathcal{F}_w,\Phi\in\mathcal{F}_{\Phi}}|\frac{1}{n_e}\langle u\circ\Phi(X^e),\bm{y}^e\rangle-\E[ \langle u\circ\Phi(\bx^{e}),y^{e}\rangle]|\leq \frac{\rho_{1,n}}{2}\right)\geq 1-c_n\nonumber\\
&\P\left(\max_{e\in\calE_{tr}}\sup_{u\in\mathcal{F}_w,\Phi\in\mathcal{F}_{\Phi}}|\frac{1}{n_e}\sum_{i=1}^{n_e}\{u\circ\Phi(\bx_i^e)\}^2-\E[\{u\circ\Phi(\bx^{e})\}^2]|\leq \frac{\rho_{2,n}}{2}\right)\geq 1-c_n\label{rho-empirical}
\end{align}
}for some $c_n=o(1)$. We see that $\rho_{1,n}$ and $\rho_{2,n}$ possibly depend on the sample size and the complexity of two functional spaces.

\begin{condition}[Diversity condition]
\label{cond1-empirical}
For $\rho_{1,n}$ and $\rho_{2,n}$ defined in (\ref{rho-empirical}), assume that  for all $\Phi\in\mathcal{F}_{\Phi}\setminus\mathcal{I}^*_{\Phi}$,
 \begin{align*}
  \text{either}~ &\max_{e,e'\in\calE_{tr}} \sup_{u\in\mathcal{F}_{w}}|\E[\langle u\circ\Phi(\bx^e),y^e\rangle]-\E[\langle u\circ\Phi(\bx^{e'}),y^{e'}\rangle]|>2\rho_{1,n}~~\nonumber\\
     \text{or}~ & \max_{e,e'\in\calE_{tr}} \sup_{u\in\mathcal{F}_{w}} |\E[\{ u\circ\Phi(\bx^e)\}^2]-\E[\{ u\circ\Phi(\bx^{e'})\}^2]|>2\rho_{2,n}.  
\end{align*}
\end{condition}
Condition \ref{cond1-empirical}  is an empirical version of the diversity condition in (\ref{cond-recover}). Under very mild conditions, both $\rho_{1,n}$ and $\rho_{2,n}$ are $o(1)$. Then Condition \ref{cond1-empirical} naturally holds under (\ref{cond-recover}) if the two quantities on the left hand-side of (\ref{cond-recover}) do not depend on $n$ for all $\Phi\in\mathcal{F}_{\Phi}\setminus\mathcal{I}^*_{\Phi}$. 
Condition \ref{cond1-empirical} can also be viewed as a signal strength condition, which ensures that for any $\Phi\in\mathcal{F}_{\Phi}\setminus\mathcal{I}^*_{\Phi}$, the true signals are sufficiently strong to surpass the noise levels $\rho_{1,n}$ and $\rho_{2,n}$ for consistent detection. 
%

\begin{theorem}[Domain generalization of empirical FAIRM]
\label{thm1-empirical}
 Suppose that Condition \ref{cond1-empirical} is satisfied and $\mathcal{I}^*_{\Phi}\neq\emptyset$. It holds that for $\rho_{1,n}$, $\rho_{2,n}$, and $c_n$ defined in (\ref{rho-empirical}), with probability at least $1-2c_n$,
$$
\max_{e\in\calE_{all}}\{R^e(\hat{w}^{(\FAIRM)}\circ\widehat{\Phi}^{(\FAIRM)})-R^e(w^*\circ\Phi^*)\}\leq 4\rho_{1,n}+2\rho_{2,n}.
$$
\end{theorem}
Theorem \ref{thm1-empirical} provides an upper bound on the domain generalization error of $\hat{w}^{(\FAIRM)}\circ\widehat{\Phi}^{(\FAIRM)}$. We see that it makes no assumptions on the true conditional mean function $\E[y^e|\bx^e]$, making it robust to model misspecification.
As defined in (\ref{rho-empirical}), $\rho_{1,n}$ and $\rho_{2,n}$ depends on the supremum over $\mathcal{F}_{\Phi}$ and $\mathcal{F}_w$. Hence, this upper bound is relatively sharp when the functional spaces are small, e.g., finite-dimensional linear functions. If the functional spaces are complex, the result in Theorem \ref{thm1-empirical} can be a loose upper bound for the domain generalization errors. Nevertheless, in certain widely-used functional spaces, such as the high-dimensional linear models considered in Section \ref{sec-lm}, the supremum can be searched based on some greedy algorithms which reduce the complexity and computational cost.

Next, we present an upper bound on the multi-calibration errors of empirical FAIRM. Let $\hat{w}_{\Phi}=\argmin_{w\in\mathcal{F}_{w}}\sum_{e\in\calE_{tr}}\alpha_e\hat{R}^e(w\circ\Phi)$ for a given representation $\Phi(\cdot)$ and
 $$\delta_n=\max_{e\in\calE_{all}}\sup_{\Phi\in\mathcal{F}_{\Phi}}\E^{1/2}[|\E[y^e|\Phi(\bx^e)]-\hat{w}_{\Phi}\circ\Phi(\bx^e)|^2].$$ 
 Here, $\delta_n$ measures the difficulty in estimating the prediction function $\E[y^e|\Phi(\bx^e)]$ based on the training data for all $\Phi\in\mathcal{F}_{\Phi}$.

\begin{theorem}[Multi-calibration of empirical FAIRM]
\label{thm2-empirical}
 Assume Condition \ref{cond1-empirical}, $\mathcal{I}^*_{\Phi}\neq \emptyset$, and $\E[y_i^e|\Phi(\bx_i^e)=\bm\phi]=w(\bm\phi)\in\calF_{w}$ for all $\Phi\in\mathcal{I}^*_{\Phi}$. Then for $\rho_{1,n}$, $\rho_{2,n}$, and $c_n$ defined in (\ref{rho-empirical}), it holds that
\begin{align*}
  &\max_{e\in\calE_{all}} \E\left[|\E[y^e|\hat{w}^{(\FAIRM)}\circ\widehat{\Phi}^{(\FAIRM)}(\bx^e)]-\hat{w}^{(\FAIRM)}\circ\widehat{\Phi}^{(\FAIRM)}(\bx^e)|\right] \\
  &\leq \sqrt{4\rho_{1,n}+2\rho_{2,n}}+C'(1+\delta_n)c_n^{1/2}.
\end{align*}
for some large enough positive constant $C'$.
\end{theorem}
 Theorem \ref{thm2-empirical} provides an upper bound on the multi-calibration error of empirical FAIRM. Typically, there exist proper tuning parameters $\rho_{1,n},\rho_{2,n}$ such that $\max\{\rho_{1,n},\rho_{2,n},c_n\}=o(1)$ in (\ref{rho-empirical}) for fixed $\mathcal{F}_w$ and $\mathcal{F}_{\Phi}$.  If $\mathcal{F}_w$ includes uniformly bounded functions, then $\delta_n=O(1)$.  Hence, the empirical FAIRM is asymptotically multi-calibrated in these cases. 

\section{FAIRM in linear models}
\label{sec-lm}

In this section, we implement FAIRM  within the scope of high-dimensional linear working models. That is, we consider invariant prediction rules with function classes ${\mathcal{F}}_{\Phi}=\{\bx_S:S\subseteq[p]\}$ and ${\mathcal{F}}_{w}=\{w: w(\bm\phi)=\bm{\phi}^\top \bm{b}\}$.  It is important to note that while the working models are linear, the true relationship between $\bx^e$ and $y^e$ remains largely unspecified, aligning with the distribution-free nature of FAIRM.

In view of (\ref{full-info}),  the full-information benchmark in linear models can be expressed as:
\begin{align}
 &\bbeta^*=\argmin_{\bm{b}\in\R^{|\Phi|}}\sum_{e\in\calE_{all}}\alpha^*_e\E[(y^e-\Phi(\bx^e)^\top \bm{b})^2],~   \text{subject to}\label{oracle-lm}\\
 ~  &\left\{S\subseteq [p]:  \max_{e,e'\in\calE_{all}}\sup_{b\in\R^{|\Phi|}}|  \E[\bm{b}^\top \bx_S^ey^e]-\E[\bm{b}^\top\bx^{e'}_Sy^{e'}]|=0\right.\nonumber\\
   &\quad\quad\quad\quad\quad \left. \max_{ e,e'\in\calE_{all}} \sup_{b\in\R^{|\Phi|}}|  \E[(\bm{b}^\top \bx^e_S)^2]-\E[(\bm{b}^\top \bx^{e'}_S)^2]|=0\right\}.\nonumber
\end{align}
We note that $\bbeta^*$ is not necessarily unique by definition (\ref{oracle-lm}), which is allowed for the purpose of prediction.
 However, directly realizing the above optimization requires searching among $2^p$ candidate values of $S$. Therefore, in the following, we propose a computationally efficient algorithm to approximate the above optimization.


\subsection{FAIRM in linear models}
\label{sec4-1}
Our proposed algorithm requires a sample splitting step. 
We denote first part of samples as $\{\mathring{X}^e,\mathring{\bm{y}}^e\}_{e\in\calE_{tr}}$ with sample size $\{\mathring{n}_e\}_{e\in\calE_{tr}}$ and the second part as $\{\tilde{X}^e,\tilde{\bm{y}}^e\}_{e\in\calE_{tr}}$ with sample size $\{\tilde{n}_e\}_{e\in\calE_{tr}}$ so that $\mathring{n}_e+\tilde{n}_e=n_e$. We will use the first part of the samples to estimate invariant prediction functions and use the second half to select the one with the best prediction performance. We require that $\min_{e\in\calE_{tr}}\mathring{n}_e\asymp \min_{e\in\calE_{tr}}n_e$ and $\sum_{e\in\calE_{tr}}\mathring{n}_e\asymp\sum_{e\in\calE_{tr}}\tilde{n}_e$ hold, which are actually easy to achieve. For instance, if $n_e$ is larger than the average group size in order, one can allocate more samples to $\tilde{n}_e$; if $n_e$ is smaller than the average group size in order, one can allocate more samples to $\mathring{n}_e$.

Let $\widehat{M}^e:=(\mathring{X}^e)^{\top}\mathring{\bm{y}}^e/\mathring{n}_e$ and $\widehat{\Sig}^e:=(\mathring{X}^e)^{\top}\mathring{X}^e/\mathring{n}_e$ denote the marginal statistics  and sample covariance matrix, respectively, based on the first fold of the $e$-th dataset. Let $\mathring{N}_{tr}:=\sum_{e\in\calE_{tr}}\mathring{n}_e$, $\mathring{\alpha}_e:=\mathring{n}_e/\mathring{N}_{tr}$, $\widehat{M}^{(tr)}_j:=\sum_{e\in\calE_{tr}}\mathring{\alpha}_e\widehat{M}^e_j$ and $\widehat{\Sig}^{(tr)}:=\sum_{e\in\calE_{tr}}\mathring{\alpha}_e\widehat{\Sig}^e$ denote the pooled statistics.
The formal algorithm is summarized in Algorithm \ref{alg1}.

\begin{algorithm}
	\SetKwInOut{Input}{Input}
	\SetKwInOut{Output}{Output}
\Input{$\{(X^e,\bm{y}^e)\}_{e\in\calE_{tr}}$, tuning parameters $\rho_M$ and $\rho_D$ as in (\ref{eq-rho}).}
\Output{Invariant prediction for $\bx^*$, $(\bx^*)^\top \hat{\bbeta}$, and estimated invariant subset $\widehat{S}$.}	

\textbf{Step 1}: Filtering $\Phi$.  For $\hat{\xi}_j^e=\sum_{i=1}^{\mathring{n}_e}(\mathring{x}_{i,j}^e\mathring{y}_i^e-
\widehat{M}^e_j)^2/\mathring{n}_e^2$, compute
\[
\widehat{I}^{(M)}=\left\{j\in[p]: \sum_{e\in\calE_{tr}}\mathring{\alpha_e}\big\{(\widehat{M}_j^e-\widehat{M}_j^{(tr)})^2-\hat{\xi}_j^e\big\}\leq \rho_M\right\}.
\]

Compute a distance matrix $\widehat{D}\in\R^{p\times p}$ such that for $\hat{\zeta}_{j,k}^e=\sum_{i=1}^{\mathring{n}_e}(\mathring{x}_{i,j}^e\mathring{x}_{i,k}^e-
\widehat{\Sig}^e_{j,k})^2/\mathring{n}_e^2$, 
\[
\widehat{D}_{j,k}=\sum_{e\in\calE_{tr}}\mathring{\alpha}_e\big\{(\widehat{\Sig}^e_{j,k}-\widehat{\Sig}^{(tr)}_{j,k})^2-\hat{\zeta}_{j,k}^e\big\}.
\]

For each $j\in \widehat{I}^{(M)}$, compute
$\widehat{C}_{j}=\{k\in \widehat{I}^{(M)}:|\widehat{D}_{j,k}|\leq \rho_D\}$.
For $j\in \widehat{I}^{(M)}$, compute
 \begin{align}
 \widehat{I}_j=\left\{\begin{array}{ll}
 \widehat{C}_j~~&\text{if}~j\in \widehat{I}^{(M)}~\text{and}~\|\widehat{D}_{\widehat{C}_j,\widehat{C}_j}\|_{\infty,\infty}\leq \rho_D\\
 \{j\}~~&\text{otherwise}.
 \end{array}\right.\label{eq-submatrix}
 \end{align}

\textbf{Step 2}: Estimating $w$.
For each $j \in \widehat{I}^{(M)}$, set $(\hat{\bbeta}^{[j]})_{\widehat{I}_j^c}=0$ and 
\begin{align*}
   (\hat{\bbeta}^{[j]})_{\widehat{I}_j}&= \argmin_{\bm{b}\in\R^{|\widehat{I}_j|}}\left\{\frac{1}{\mathring{N}_{tr}}\sum_{e\in\calE_{tr}}\|\mathring{\bm{y}}^e-\mathring{X}_{.,\widehat{I}_j}^e\bm{b}\|_2^2+\lambda_j \|\bm{b}\|_1\right\}.
\end{align*}

\textbf{Step 3}: Find optimal solutions with minimum training errors based on some independent samples $\{\widetilde{X}^e,\tilde{\bm{y}}^e\}_{e\in\calE_{tr}}$.
\begin{align}
    \label{eq-that}
\hat{j}=\argmin_{j\in\widehat{I}^{(M)}}\sum_{e\in\calE_{tr}}\|\tilde{\bm{y}}^e-\widetilde{X}^e\hat{\bbeta}^{[j]}\|_2^2.
\end{align}

Set $\widehat{S}=\widehat{I}_{\hat{j}},~~\hat{\bbeta}=\hat{\bbeta}^{[\hat{j}]}$.
\caption{FAIRM in linear models}
\label{alg1}
\end{algorithm}

 In Step 1, we screen out covariates that have large variability in the marginal statistic $\widehat{M}^e_j$. The quantity $\hat{\xi}^e_j$ adjusts the finite-sample bias in the first term. This step is a finite-sample realization of the first constraint in FAIRM. Next, we measure the variability in $\{\widehat{\Sig}^e\}_{e\in\calE_{tr}}$ by $\widehat{D}$ and the quantity $\hat{\zeta}^e_{j,k}$ corrects the finite-sample bias. 
For each $j\in\widehat{I}^{(M)}$, we find $\widehat{C}_j$ such that the variability of $\widehat{\Sig}_{j,\widehat{C}_j}^e$ for $e\in\mathcal{E}_{tr}$, measured by $\widehat{D}_{j,\widehat{C}_j}$, is small enough. If $\widehat{D}_{\widehat{C}_j,\widehat{C}_j}$ is small enough, then $\widehat{C}_j$ satisfies both constraints and we include it as a candidate invariant representation, denoted by $\widehat{I}_j$.  We see that the optimization (\ref{eq-submatrix}) reduces traversing over $2^{|\widehat{I}^{(M)}|}$ representations to $|\widehat{I}^{(M)}|$ ones, which significantly reduces the computational cost. We will show later that $\widehat{I}_j\in\mathcal{I}^{*}_{\Phi}$ with high probability for each $j\in \widehat{I}^{(M)}$.

 In Step 2, we compute the prediction rule based on each candidate invariant representation. The $\ell_1$-penalty is to increase the estimation accuracy when $|\widehat{I}_j|$ is relatively large. In Step 3, we search for the invariant prediction rule which has the smallest training errors corresponding to the objective function of FAIRM. This step is based on independent samples in the same spirit of cross-validation.

\subsection{Theoretical properties}
\label{sec4-2}
In this subsection, we establish theoretical guarantees for Algorithm \ref{alg1}.

\begin{condition}[Model assumptions]
\label{cond-model}
For each $e\in\calE_{tr}$, assume that $y_i^e$ are independent sub-Gaussian with mean zero and $\max_{e\in\calE_{tr}}\E[(y_i^e)^2]\leq c_1<\infty$. 
For each $e\in\calE_{tr}$, assume that $\bx_{i}^e$ are independent sub-Gaussian vectors with mean zero and covariance matrix $\Sig^e$. We assume that $c_2\leq \min_{e\in\calE_{tr}}\Lambda_{\min}(\Sig^e)\leq \max_{e\in\calE_{all}}\Lambda_{\max}(\Sig^e)\leq c_3$ and $\Sig^e_{j,j}=\Sig^{e'}_{j,j}$ for all $e,e'\in\calE_{tr}$, $j=1,\dots,p$. 
\end{condition}
Condition \ref{cond-model} assumes sub-Gaussian observations. Note that we do not assume $\E[y_i^e|\bx_i^e]$ is linear but rather consider linear prediction functions as working models.
The sub-Gaussian distributions allow us to choose tuning parameters such that
{\small
\begin{align}
\rho_M=C_0\frac{(\log p)^2+\sqrt{|\calE_{tr}|\log p}}{N_{tr}}~\text{and}~~\rho_D=C_0\frac{(\log p)^2+\sqrt{|\calE_{tr}|\log p}}{N_{tr}},\label{eq-rho}
\end{align}
}
where $C_0$ is a proper constant and $N_{tr}=\sum_{e\in\calE_{tr}}n_e$. 

Similar to the diversity assumption in Proposition~\ref{prop-recover}, we need a condition that guarantees the variant features have sufficiently large variability so that they will be screened out with high probability. Let $M^e=\E[\widehat{M}^e]$, $M^{(tr)}=\E[\widehat{M}^{(tr)}]$, $\Sig^{(tr)}=\E[\widehat{\Sig}^{(tr)}]$, and $I^{(M)}=\{j\leq p: \max_{e\in\calE_{all}} |M^e_j-M^{e'}_j|=0\}$.
Define $\mathcal{S}^*=\{S\subseteq[p]:\bx_S\in\mathcal{I}^*_{\Phi}\}$, which is the collection of subsets delivering invariant representations. We note that there is a one-to-one correspondence between $\mathcal{S}^*$ and $\mathcal{I}^*_{\Phi}$.

\begin{condition}[Significant variability for variant features]
\label{cond2}
For any $S\notin\mathcal{S}^*$, either 
\[
 \|\sum_{e\in\calE_{tr}}\alpha_e(M^e_S-M^{(tr)}_S)^2\|_{\infty}\geq 2\rho_M~\text{or}~\|\sum_{e\in\calE_{tr}}\alpha_e(\Sig^e_{S,S}-\Sig^{(tr)}_{S,S})^2\|_{\infty,\infty}>2\rho_D.
 \]
\end{condition}
Condition \ref{cond2} can be viewed as a minimum signal strength condition for consistently detecting variant representations. As specified in (\ref{eq-rho}), $\rho_M$ and $\rho_D$ go to zero as the training sample sizes get larger and hence Condition \ref{cond2} gets weaker when the sample sizes grow to infinity. Condition \ref{cond2} also gets weaker when $\calE_{tr}$ gets larger, i.e., when the training data get more representative.

\begin{lemma}[Invariance of the selected features]
\label{lem-select1}
Assume Condition \ref{cond-model} and Condition \ref{cond2}. Given that $\log p\lesssim \min_{e\in\calE_{tr}}n_e$, there exists some positive constant $c_1$ such that
\[
\P\left(\widehat{I}_j\in\mathcal{S}^{*},~\forall j\in\widehat{I}^{(M)}\right)\geq 1-\exp\{-c_1\log p\}.
\]
\end{lemma}

Lemma \ref{lem-select1} states that with high probability, $\phi(\bx)=\bx_{\widehat{I}_j}$, $j\in\widehat{I}^{(M)}$, are all feasible invariant representations to the full-info FAIRM (\ref{oracle-lm}) and hence they are all invariant representations by Proposition \ref{prop1}. 

Next, we show that $\hat{\bbeta}$, which is computed based on the selected $\widehat{I}_{\hat{j}}$, has comparable performance as full-info FAIRM in linear models. Denote the optimal feasible solutions to (\ref{oracle-lm}) as
    \[
    \{S^*_1,\dots, S^*_K\}=\argmax_{S\in\mathcal{S}^*} \sum_{e\in\calE_{all}} \alpha^*_e(M^e_S)^\top \{\Sig^e_{S,S}\}^{-1}M^e_S.
    \]
We need one more condition to ensure that at least one $S_k^*$ is selected as one of $\{\widehat{I}_j\}_{j\in\widehat{I}^{(M)}}$ with high probability. 
\begin{definition}[Anchor index]
For a set $S\in\mathcal{S}^{*}$, we say that $S$ has an anchor index $j$ if $j\notin S'$ for any $S'\in\mathcal{S}^*$ and $S'\not\subseteq S$. 
\end{definition}

    \begin{condition}[Condition on the optimal solutions]
    \label{cond-anchor}
 Assume that there exists $S^*\in\{S^*_1,\dots, S^*_K\}$ such that    (a) $S^*$ has an anchor index $j^*$;
(b) $\|\bbeta^*\|_0\log p=o(N_{tr})$, where $\bbeta^*\in\R^p$ is such that\\
 $\bbeta^*_{S^*}=\{\sum_{e\in\calE_{all}}\alpha^*_e\Sig^e_{S^*,S^*}\}^{-1}\{\sum_{e\in\calE_{all}}\alpha^*_eM_{S^*}^e\}$ and $\bbeta^*_j=0$ for all $j\notin S^*$.
    \end{condition}
 Condition \ref{cond-anchor} (a) requires at least one of the optimal solutions $\{S_1^*,\dots,S_K^*\}$ to have an anchor index. For instance, in Example \ref{ex1}, we can check that Condition \ref{cond-anchor} (a) is satisfied as $S^*=\{1,2\}$, which has the largest prediction power. This condition also naturally holds if at least one $S^*_k$ is exclusive of other sets in $\mathcal{S}^*$. 
Compared to the existing literature, \citet{fan2023environment} requires a unique invariant set under their definitions, which is stronger than Condition \ref{cond-anchor}; see the discussion under their Condition 5. \citet{peters2016causal} does not require similar conditions as they only aim to identify the intersection of all the invariant subsets, which can have low prediction power as we discussed in Section \ref{sec2-1}.

\begin{theorem}[Generalization bounds of Algorithm \ref{alg1}]
\label{thm1}
Assume Conditions \ref{cond-model}, \ref{cond2}, and \ref{cond-anchor}. Then for $\lam_j=c_1\sqrt{\log p/N_{tr}}$ with some large enough constant $c_1$, we have
\begin{align}
&\P\left(\exists j\in[p], ~S^*=\widehat{I}_{j}\right)\geq 1-\exp\{-c_2\log p\}\label{eq1-thm1}\\
&\max_{e\in\calE_{all}}\left\{R^e(\hat{\bbeta})-R^e(\bbeta^*)\right\}\lesssim \frac{\|\bbeta^*\|_0\log p}{N_{tr}}+\frac{ t}{\sqrt{N_{tr}}}\label{eq2-thm1}
\end{align}
with probability at least $1-\exp\{-c_1\log p\}-\exp\{-c_2t^2\}$.
\end{theorem}
The expression in (\ref{eq1-thm1}) states that with high probability,  at least one optimal invariant subset $S^*$ is among the selected subsets $\{\widehat{I}_j\}_{j\in\widehat{I}^{(M)}}$. In the proof, we essentially show that if an invariant subset $S^*$ has an anchor index $j^*$, then $\widehat{I}_{j^*}=S^*$ with high probability.

In (\ref{eq2-thm1}), we show that $\hat{\bbeta}$ has prediction performance comparable to the full-info FAIRM in linear models, no 
matter which environments the test samples are taken from. We mention that the prediction error bound (\ref{eq2-thm1}) does not imply an estimation error bound of $\hat{\bbeta}$ because $\bbeta^*$ may not be unique. We prove the minimax optimality of $\hat{\bbeta}$ in terms of prediction accuracy in Theorem \ref{thm-minimax}. We also note that the results of Theorem \ref{thm1} do not rely on the linearity of $\E[y^e|\bx^e]$ and hence it is distribution-free.

In the next corollary, we investigate the fairness guarantee of the prediction function $\bm x^\top \hat{\bbeta}$.  
\begin{corollary}[Fairness guarantee of Algorithm \ref{alg1}]
\label{cor-mc}
Assume the Conditions of Theorem \ref{thm1}. 
Assume that $\E[y^e|\bx^e]=(\bx^e)^\top \bbeta^e$ for some $\bbeta^e\in\R^p$ and $\bx^e$ are Gaussian distributed $\forall e\in\calE_{all}$.
Then
\[
\sup_{e\in\mathcal{E}_{all}}\E\left[|\E[y^e|(\bx^e)^\top \hat{\bbeta}]-(\bx^e)^\top \hat{\bbeta}|\right]\leq c_2\sqrt{\frac{\|\bbeta^*\|_0\log p}{N_{tr}}}+c_3\left(\frac{\log p}{N_{tr}}\right)^{1/4}
\]
for some positive constants $c_2$ and $c_3$.
\end{corollary}
Corollary \ref{cor-mc} is a direct consequence of Theorem \ref{thm1} and the generic upper bound in Theorem \ref{thm2-empirical}. The Gaussian assumption on the design guarantees that $\E[y_i^e|\bx_{i,S}^e]$ is linear for all $S\subseteq [p]$. A similar condition is needed in justifying the fairness of full-info FAIRM in Proposition \ref{prop1}.

\subsection{Minimax lower bound}

In this subsection, we establish the minimax lower bound for estimating the invariant prediction functions.
Define our parameter space of interest as
\begin{align*}
  \Xi(s,\rho,\sigma^2)&=\left\{\mathcal{P}=\{P^e\}_{e\in\calE_{all}}: \max_{e\in\calE_{all}}\E[\{\E[y^e|\bx^e]-(\bx^e)^\top \bbeta^*\}^2]\leq \rho,\right.\\
  &\quad\quad\left.\max_{e\in\calE_{all}}\text{Var}(y^e|\bx^e)\leq \sigma^2,~\text{(\ref{cond-recover}) holds},~\|\bbeta^*\|_0\leq s\right\}.
\end{align*}
We define the loss function as:
\begin{align*}
r(\hat{\bbeta},\mathcal{P})=
\left\{\begin{array}{ll}
\max\limits_{e\in\calE_{all}}\left\{R^e(\hat{\bbeta})-R^e(\bbeta^*)\right\}~&\text{if}~supp(\hat{\bbeta})\in\mathcal{I}^{*}_{\Phi}\\
\infty&~\text{otherwise}.
\end{array}
\right.
\end{align*}
To be more precise, in the above definitions, the parameters $\bbeta^*=\bbeta^*(\mathcal{P})$ and $\mathcal{I}^*_{\Phi}=\mathcal{I}^*_{\Phi}(\mathcal{P})$ are with respect to $\mathcal{P}=\{P^e\}_{e\in\calE_{all}}$ as defined in (\ref{oracle-lm}). This loss function focuses on the excess risk within invariant prediction rules, assigning a higher loss to any prediction rule not satisfying the invariance criterion.
\begin{theorem}[Minimax lower bound for invariant prediction errors]
\label{thm-minimax}
Assume Condition \ref{cond-model} holds for $\mathcal{P}\in\Xi(s,\rho,\sigma^2)$. For $3\leq s\leq c_1N_{tr}/\log p$ and $\rho\geq 1$, it holds that
\begin{align*}
\inf_{\hat{\bbeta}}\sup_{\mathcal{P}\in\Xi(s,\rho,\sigma^2)}\E_{\mathcal{P}}[r(\hat{\bbeta},\mathcal{P})]\geq C\frac{\sigma^2s\log p}{N_{tr}}+\frac{C}{\sqrt{N_{tr}}}.
\end{align*}
\end{theorem}
Theorem \ref{thm-minimax} establishes the minimax lower bound for $r(\hat{\bbeta},\mathcal{P})$, which concerns the minimax excess risk within the class of invariant predictors. Comparing with the upper bound in Theorem \ref{thm1}, we see that the proposed prediction rule $\bm{x}^\top \hat{\bbeta}$ achieves minimax optimal error rate.

\section{Numerical Studies}
\label{sec-simu}
In this section, we conduct two numerical experiments with synthetic data and two additional experiments based on the MNIST data. The code for all the experiments are available at \url{https://github.com/saili0103/FAIRM/}.
\subsection{Prediction experiment: Identify covariance matrix}
We set  the number of features  $p=400$, training environments $\calE_{tr}=\{1,\dots,K\}$ with $K$ ranging from 4 to 12, test environments $\calE_{te}=\{K+1,K+2,K+3\}$, and $\calE_{all}=\calE_{tr}\cup\calE_{te}$. Let $S^*=\{1,\dots,10\}$ denote the set of invariant predictors with nonzero effects and $\beta_j=1-0.15j$ for $j=1,\dots,10$. A subset $S_v$ is uniformly random selected in $[p]\setminus S^*$ represents the set of variant features with $|S_v|=5$. 
For each $e\in\calE_{all}$, data is generated as follows. For $i=1,\dots,n_e=100$,
\begin{align*}
&\bm{z}^e_i\sim N(0,\Sig_z),~x^e_{i,j}=z^e_{i,j}~\text{if}~j\notin S_v,\\
&y_i^e=(\bx^e_{i,S^*})^\top \bbeta_{S^*}+ N(0,1),~x_{i,j}^e=z_{i,j}^e+y_i^e\gamma_j^e~\text{if}~j\in S_v,
\end{align*}
where $\Sig_z\in\R^{p\times p}$ will be specified later and $\gamma^e_j\stackrel{i.i.d.}{\sim} N(0,\delta^2)$ determines the level of confounding. We see that all the features in $[p]\setminus S_v$ are within $\mathcal{I}^*_{\Phi}$. We set $n_e=100$ for all $e\in\calE_{tr}$.

To evaluate the performance of a prediction function $f(\cdot)$, we generate independent data $\{\check{X}^e,\check{\bm{y}}^e\}_{e\in\calE_{all}}$ with sample size $\check{n}_e$. For the error metrics, we consider the largest prediction error across all the environments $\max_{e\in\calE_{all}}\|\check{\bm{y}}^e-f(\check{X}^e)\|_2^2/\check{n}_e$ (called domain generalization errors in the following discussions). For algorithmic fairness, we consider the multi-calibration error
$\max_{e\in\calE_{all}}|\bm{1}^\top (\check{\bm{y}}^e-f(\check{X}^e))|/\check{n}_e$.  

\begin{figure}
\centering
\includegraphics[height=5.6cm, width=0.49\textwidth]{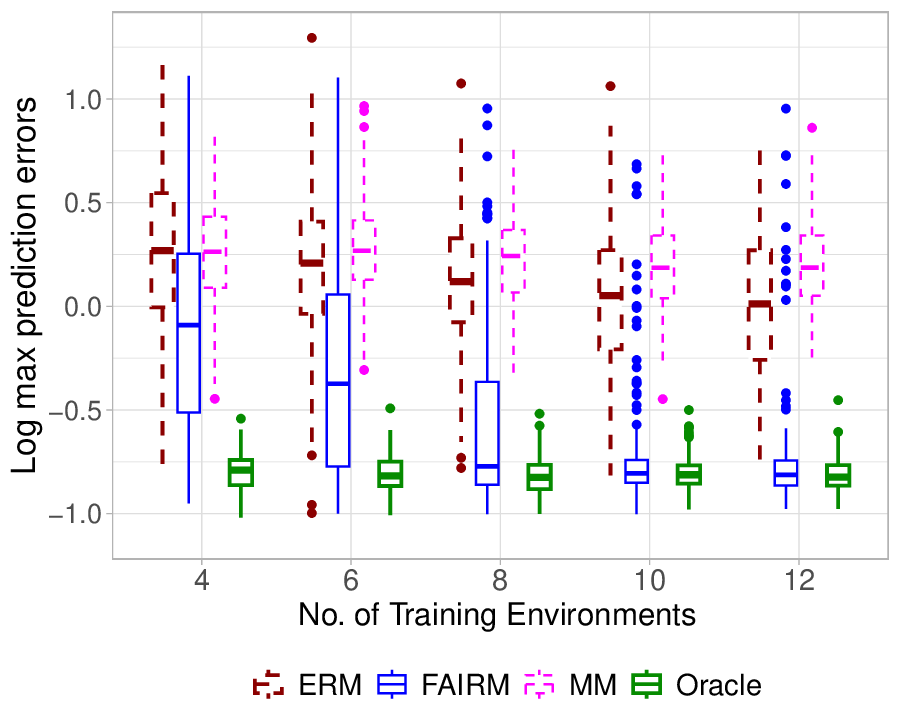}
\includegraphics[height=5.6cm, width=0.49\textwidth]{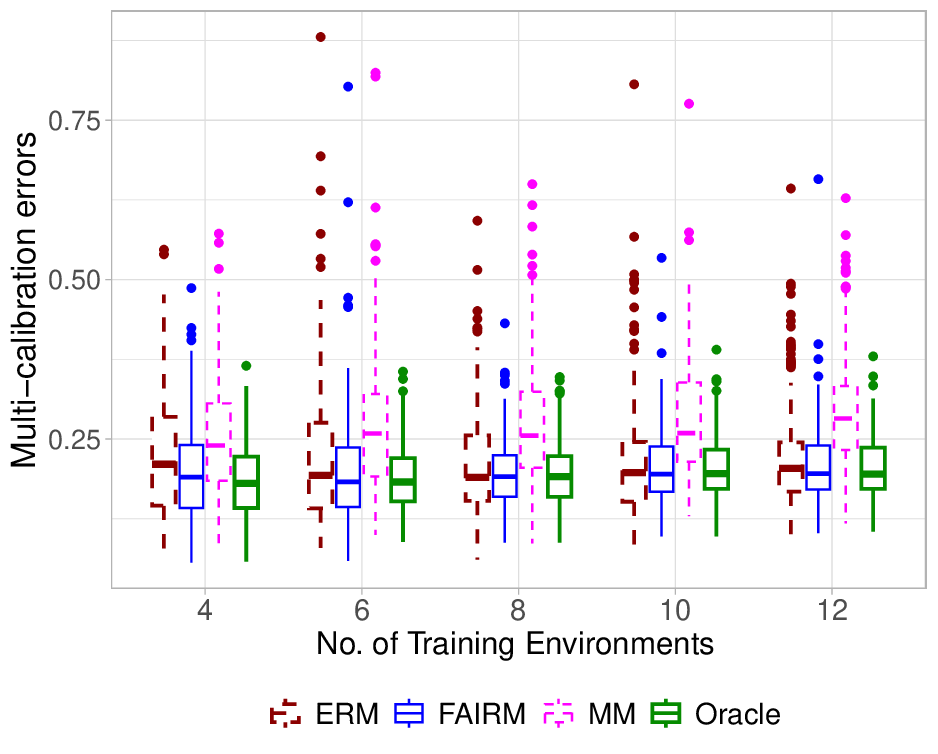}
\caption{Performance of four methods with $\Sig_z=I_p$ and $\delta=0.6$. ``ERM'' denotes the single-task lasso based on the training data. ``FAIRM'' denotes Algorithm \ref{alg1}. ``MM'' denotes the Maximin method based on the training data. The ``Oracle'' method is the single-task Lasso based on $\{X^e_{.,S_v^c},\bm{y}^e\}_{e\in\calE_{tr}}$. Each boxplot is based on 200 independent replications.}
\label{fig-dg1}
\end{figure}

In the first experiment, we set $\Sig_z=I_p$ and $\delta=0.6$. In Figure \ref{fig-dg1}, we see that as $|\calE_{tr}|$ increases, the max prediction error of FAIRM gets smaller and closer to the oracle setting. This is because  as $|\calE_{tr}|$ increases, the training environments get more diverse and $N_{tr}$ gets larger, which make it easier to distinguish variant features from invariant ones.
The two other methods have larger prediction errors in the test domains, which aligns with our analysis in Section \ref{sec3-2}. For the multi-calibration errors, we see a similar decrease for FAIRM as the number of training environments grows. Two other methods have relatively larger calibration errors.

\begin{figure}
\includegraphics[height=5.6cm, width=0.48\textwidth]{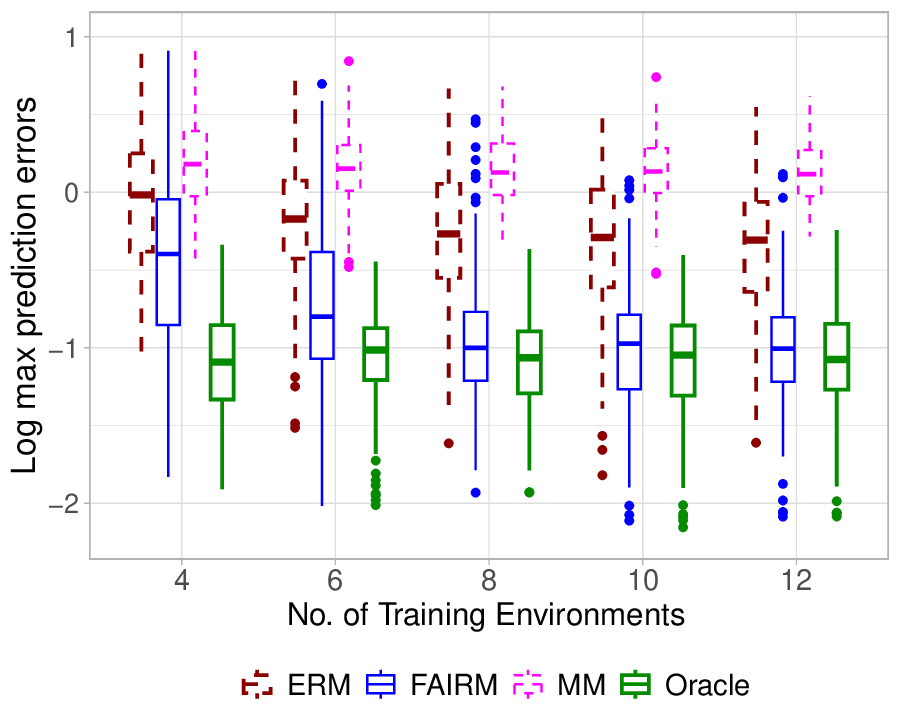}
\includegraphics[height=5.6cm, width=0.48\textwidth]{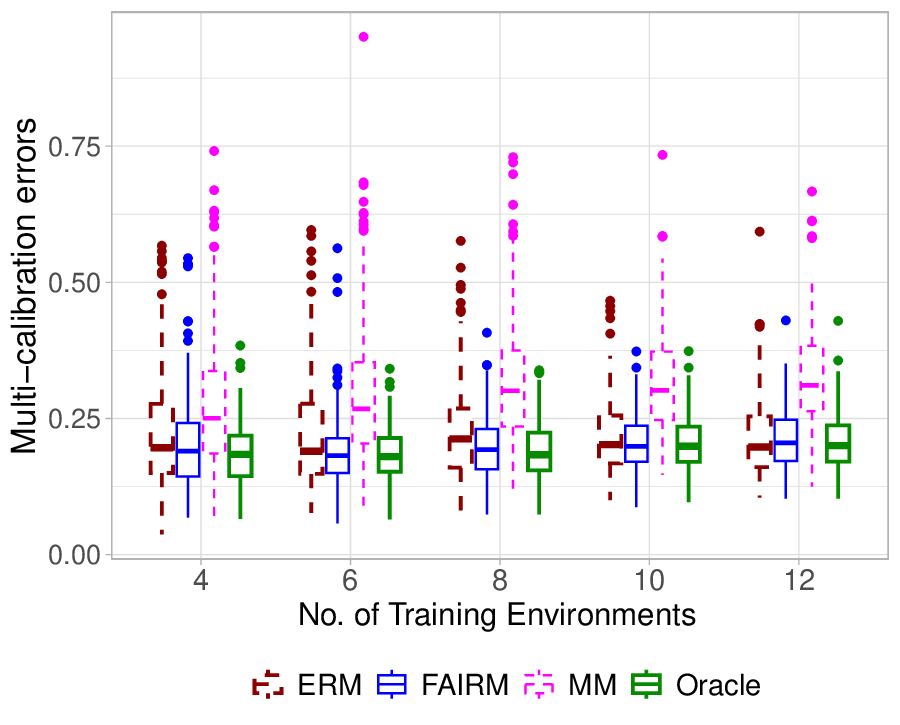}
\caption{Performance of four methods with equi-correlated $\Sig_z$ and $\delta=0.4$. ``ERM'' denotes the single-task lasso based on the training data. ``FAIRM'' denotes Algorithm \ref{alg1}. ``MM'' denotes the Maximin method based on the training data. The ``Oracle'' method is the single-task Lasso based on $\{X^e_{.,S_v^c},\bm{y}^e\}_{e\in\calE_{tr}}$. Each boxplot is based on 200 independent replications.}
\label{fig-dg2}
\end{figure}

In the second experiment, we set $\delta=0.4$ and $\Sig^z_{j,k}=0.7+0.3\mathbbm{1}(j=k)$ which gives an equi-correlated covariance matrix of $\Sig^z$.  The results, as shown in Figure \ref{fig-dg2}, display a similar pattern to the first experiment, confirming FAIRM's reliability with correlated covariates. In the supplement (Section D), we also report the training error $\sum_{e\in\calE_{tr}}\|\bm{y}^e-f(X^e)\|_2^2/N_{tr}$ and prediction performance on the unseen domains $\max_{e\in\calE_{te}}\|\check{\bm{y}}^e-f(\check{X}^e)\|_2^2/\check{n}_e$, which shows that while ERM has smaller training errors but FAIRM outperforms ERM in generalization in the unseen test data.

\subsection{Color MNIST}
This section evaluates the proposed method using a setup similar to the Color MNIST experiment \citep{rojas2018invariant}, which is a classification task based on the MNIST dataset (\url{http://yann.lecun.com/exdb/mnist/}).

\begin{figure}
\centering
\includegraphics[height=8cm,width=0.9\textwidth]{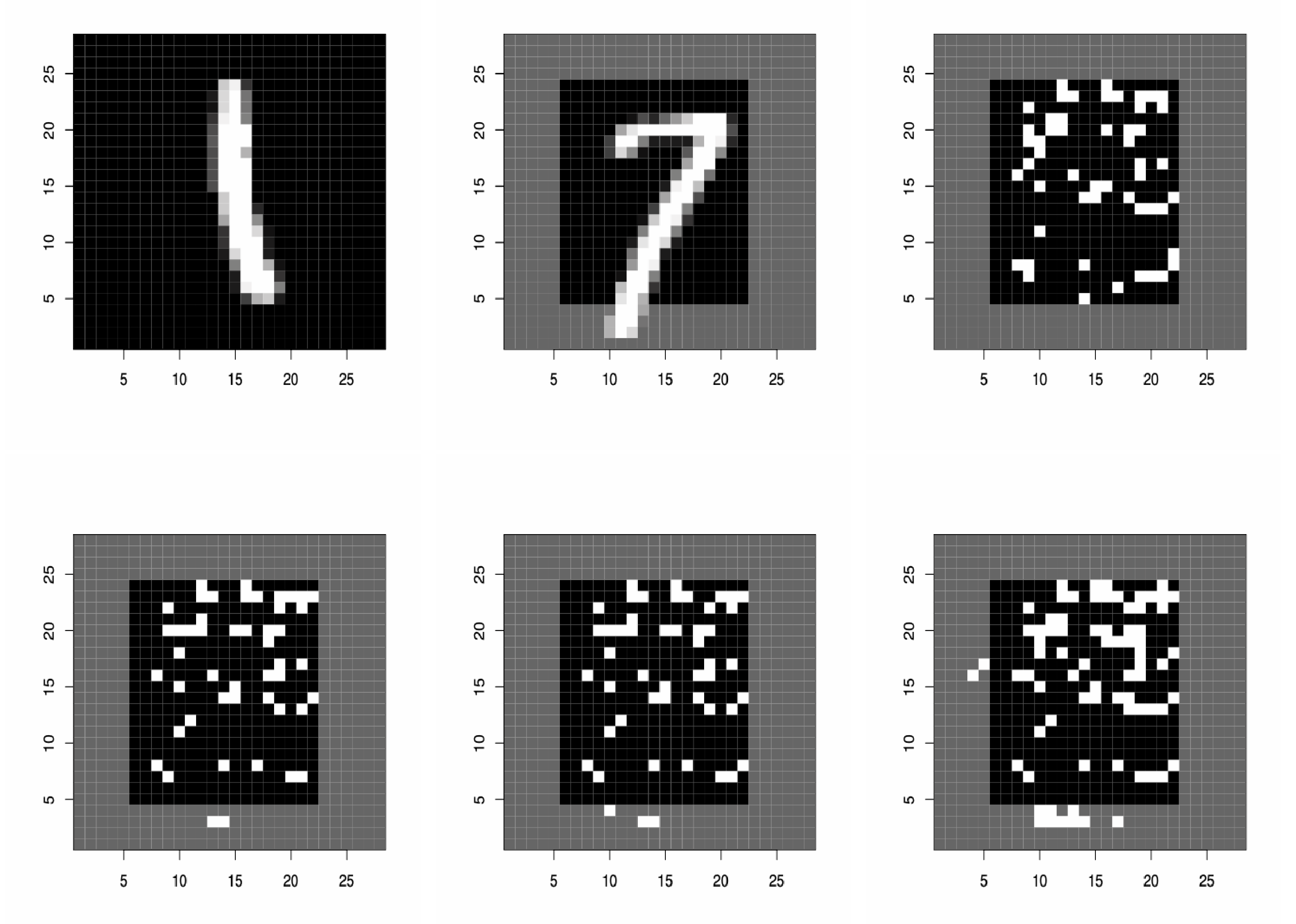}

\caption{Left and middle top: samples with different labels can have different frame colors, showing that the labels and frame colors are spuriously correlated. Right top: The oracle invariant set excluding frame features  (black) and the estimated nonzero correlations within the oracle set (white). Bottom: The estimated nonzero correlations given by FAIRM (left), ERM (middle), and Maximin (right).}
\label{fig1-mnist}
\end{figure}

In the first experiment, we use 1000 training images, randomly sampled from 13007 images whose labels are either 1 or 7. Each image has size $28\times 28$ which leads to $p=756$ features. For the $i$-th image, let $\bx_i\in\R^{756}$ denote the grayscale at each location and $y_i=1$ if the label is $7$ and $y_i=0$ if the label is 1.
We give each image a ``frame'' such that the frame color correlates (spuriously) with the class label. As in Figure \ref{fig1-mnist}, the ``frame'' is defined as multiple rows and columns surrounding the image.
By construction, any algorithm purely minimizing training error will tend to exploit the colored frame. 

 We define training environments $\calE_{tr}=\{1,2\}$ and test environments $\calE_{te}=\{3,4,5\}$.  For each environment, we set $y^e_i=y_i$ for all $e\in\calE_{all}$ and generate $\bx^e_i$ as follows. We first generate $b_i^e$ from a Bernoulli distribution with probability $p_e$ for each $e\in\calE_{all}$. Let $z_i^e=b_i^ey_i+(1-b_i^e)(1-y_i)$. That is, $z_i^e=y_i$ with probability $p_e$. We set $x_{i,j}^e=100z_i^e$ if $x^e_{i,j}$ is located in the frame. Hence, the grayscale of the frame is correlated with the outcome and the strength of correlation is determined by $p_e$. It is easy to show that if $p_e>0.5$, the grayscale of frame is positively correlated with $y_i^e$ and if $p_e<0.5$, the correlation is negative.
We set $p_1=0.9$ and $p_2=0.6$ for the training environments, and $p_3=0.8$, $p_4=0.5$, and $p_5=0.2$ for the test environments. We note that the frame color is positively correlated with the outcome in both training environments.

In the second row of Figure \ref{fig1-mnist}, the white squares of each plot denote the support of the estimated coefficients 
and the white squares in the frame region denote spurious correlations. We see that the Maximin estimator tends to exploit the color most significantly and our proposal is the opposite. 
From the results for experiment 1 in Table \ref{tab1}, we see that FAIRM, ERM, and Maximin have test errors increase as $p_e$ decreases but the errors of FAIRM increase mildly. This is because if the prediction rules exploit the color in the training phase and it learns the positive correlation then it can have large prediction errors in the test domain with $p_e<0.5$.  FAIRM also has the smallest domain generalization errors among the three methods. The oracle prediction rule has the prediction error unchanged in different domains as it does not exploit the colors at all. For the multi-calibration errors, we observe a similar pattern as those for the prediction errors.

In the second task, we consider classifying images with labels 0 or 6. The training and test environments are generated the same as in the first task but with $p_2=0.4$. Distinguished from the first task, this set-up ensures the frame colors are negatively correlated with the outcome in the second environment.
We see that the proposed method successfully recovers the oracle invariant set and has the best prediction accuracy. The Maximin estimator has smaller domain generalization errors than ERM but large calibration errors.

\begin{figure}
\centering
\includegraphics[height=8cm,width=0.9\textwidth]{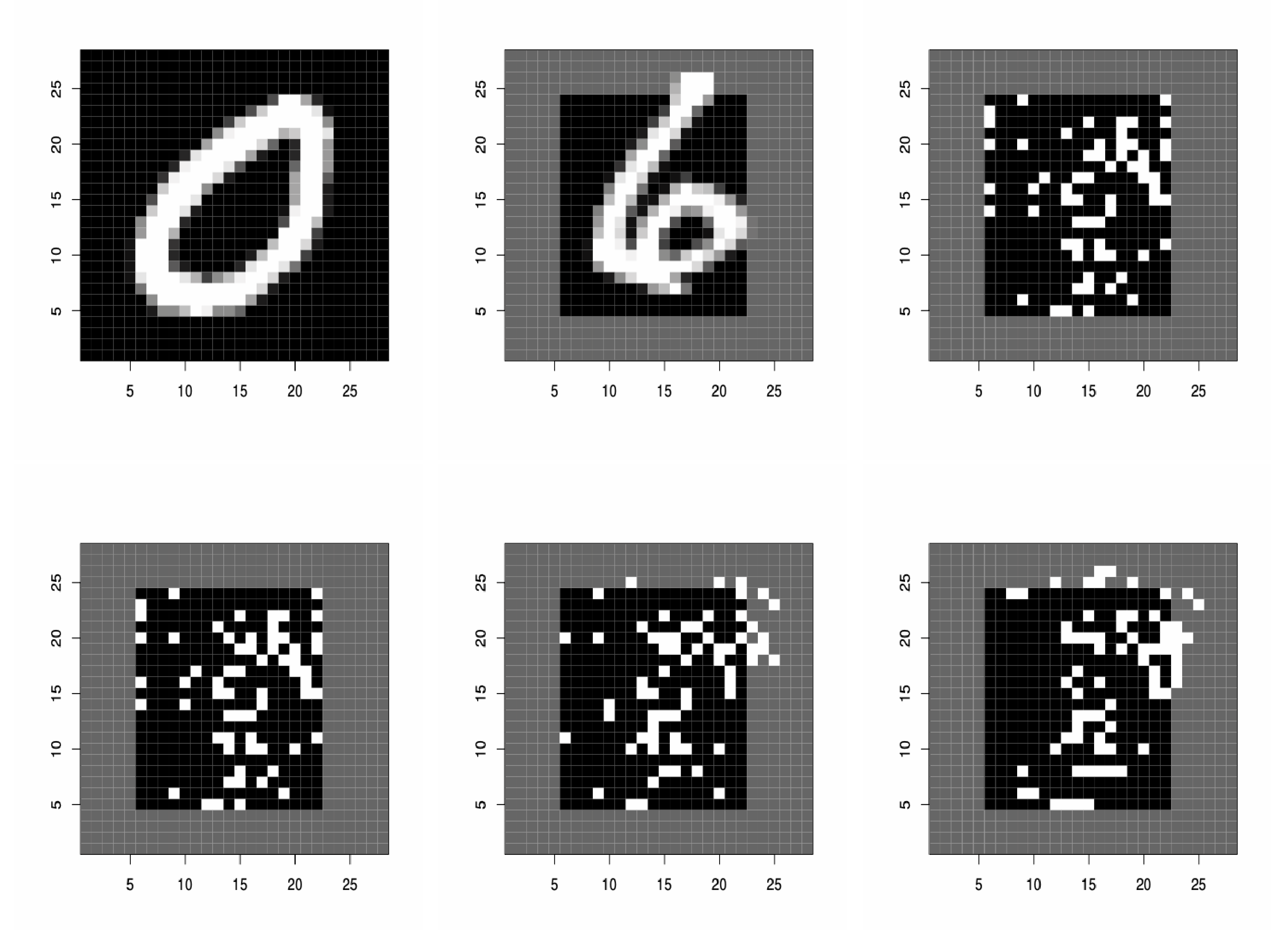}
\caption{\label{fig2-mnist} Left and middle top: samples with different labels can have different frame colors, showing that the labels and frame colors are spuriously correlated. Right top: The oracle invariant set excluding frame features  (black) and the estimated nonzero correlations within the oracle set (white). Bottom: The estimated nonzero correlations given by FAIRM (left), ERM (middle), and Maximin (right).}
\end{figure}

\begin{table}

\caption{\label{tab1} Training (Train) and test classification errors (Test) for two Color MNIST tasks in three test domains with different $p_e$.}

\begin{tabular}{|c|c|c|ccc|ccc|}
\hline
Task &Method& Train & Test(0.8) & Test(0.5) & Test(0.2) \\
 \hline
 \multirow{3}{*}{1 vs 7}&FAIRM  &0.0006 & 0.0116 & 0.0125 & 0.0148\\
\cline{2-6}
&ERM& 0.0006 & 0.0116 & 0.0125 & 0.0153 \\
\cline{2-6}
&Maximin & 0.0046  & 0.0111 & 0.0166 & 0.0236 \\
\cline{2-6}
&Oracle &0.0007& 0.0134 & 0.0134 & 0.0134 \\
\hline
\multirow{3}{*}{0 vs 6}&FAIRM  &0.0007& 0.0098 & 0.0098 & 0.0098  \\
\cline{2-6}
&ERM&0.0006 & 0.0072 & 0.0098 & 0.0119 \\
\cline{2-6}
&Maximin & 0.0106& 0.0114 & 0.0108 & 0.0103 \\
\cline{2-6}
&Oracle & 0.0011& 0.0098 & 0.0098 & 0.0098 \\
\hline
\end{tabular}

\end{table}

\section{Discussion}
\label{sec-diss}
This work studies a new invariant prediction framework, FAIRM, and its empirical realizations. We prove that FAIRM has desirable domain generalization and fairness properties in comparison to existing methods. It can be easily adapted to linear models which leads to a distribution-free algorithm with minimax optimality.

We conclude the paper by discussing the applications of the FAIRM framework to other model classes. The first extension is nonlinear prediction functions, which can be specified by $\mathcal{F}_{\Phi}=\{\bx_S:S\subseteq[p]\}$ and a nonlinear function class $\mathcal{F}_{w}=\{w: w\in L_2\}$, where $L_2$ denotes the class of Lipschitz functions. This leads to a class of single-index models which can capture non-linear relationships. 
Another widely used class of functions is 
 $\mathcal{F}_{\Phi}=\{\bx^\top V: V\in{\mathcal O}^{p\times r} \}$ where $\mathcal O^{p\times r} $ is the class of $p$ by $r$ orthonormal matrices.
 This type of linear representation of features has been considered in recent machine learning literature of representation learning \citep{du2020few, tripuraneni2021provable, lee2021predicting}. 
 It is also of interest to develop corresponding theory of learning such invariant representations as a future research direction.


\section{Acknowledgement}
The authors would like to thank Cynthia Dwork for her thoughtful comments and insights that helped us improve the paper.
\bibliography{cite}
\bibliographystyle{chicago}

\end{document}